# Kernel Density Feature Points Estimator for Content-Based Image Retrieval

Tranos Zuva[1], Oludayo O. Olugbara[2], Sunday O. Ojo[3] and Seleman M. Ngwira[4]

[1, 4]Department of Computer Engineering, Tshwane University of Technology, Pretoria, South Africa, `zuvat@tut.ac.za`

[2]Department of Information Technology, Durban University of Technology, Durban, South Africa

[3]Faculty of Information and Communication Technology, Tshwane University of Technology, Pretoria, South Africa

## ABSTRACT

*Research is taking place to find effective algorithms for content-based image representation and description. There is a substantial amount of algorithms available that use visual features (color, shape, texture). Shape feature has attracted much attention from researchers that there are many shape representation and description algorithms in literature. These shape image representation and description algorithms are usually not application independent or robust, making them undesirable for generic shape description. This paper presents an object shape representation using Kernel Density Feature Points Estimator (KDFPE). In this method, the density of feature points within defined rings around the centroid of the image is obtained. The KDFPE is then applied to the vector of the image. KDFPE is invariant to translation, scale and rotation. This method of image representation shows improved retrieval rate when compared to Density Histogram Feature Points (DHFP) method. Analytic analysis is done to justify our method, which was compared with the DHFP to prove its robustness.*

## 1. INTRODUCTION

The vast collection of digital images on personal computers, institutional computers and Internet necessitates the need to find a particular image or a collection of images of interest. This has motivated many researchers to find efficient, effective and accurate algorithms that are domain independent for representation, description and retrieval of images of interest. There have been many algorithms developed to represent, describe and retrieve images using their visual features such as shape, colour and texture [1], [2], [3], [4], [5]. The visual feature representation and description play an important role in image classification, recognition and retrieval. A successful image representation and description is dependent on the selection of suitable image features to encode and quantify these features [4].

Shape representation and description have been dominant in research area of image processing because shape is considered to be the basis of human visual recognition [4]. The shape representation can be classified as region based or contour based. The contour based techniques use the boundary of shape to describe an object. It is commonly believed that human beings can differentiate objects by their boundaries or contours [2]. Usually, most objects form shapes with defined contours, making the use of these techniques most appealing. The techniques can generally be applied to different application areas with a considerable success. The techniques have a low computation complexity as compared to region based techniques and they are sensitive to noise. These techniques in this group are well described in [5].

The region based shape representation uses the boundary pixels and the interior pixels of the shape. This group of shape representation algorithms are robust to noise, shape distortion and they are applicable to generic shapes [6]. These techniques can be found in [5]. This paper proposes a Kernel Density Feature Points Estimator (KDFPE) representation of an image object. This method imitates human visualization of image object shape and matching similar object shapes. A comparison of retrieval of similar image object shapes is done between KDFPE and Density Histogram of Feature Points (DHFP) representation of image object shapes.

## 2. SHAPE REPRESENTATION BY KDFPE

This method describes the feature points within rings in an image grid. Assume we have a silhouette object shape segmented by some means such as active contour without edges [7] and let the feature points set $P(x,y)$ (intensity function) of the object shape be defined as

$$P(x,y) = p_i(x,y), i = 1, 2, \ldots n, n \varepsilon \mathrm{N}. \tag{1}$$

The centroid of the object shape is calculated. The following formulae will be used to calculate the centroid [8],[9]:

$$x_c = \frac{m_{1,0}}{m_{0,0}} \tag{2}$$

$$y_c = \frac{m_{0,1}}{m_{0,0}} \tag{3}$$

where $m_{1,0}, m_{0,1}, m_{0,0}$ are derived from the silhouette moments given by

$$m_{i,j} = \sum_x \sum_y x^i y^j P(x,y). \tag{4}$$

The theorems that guarantee the uniqueness and existence of silhouette moments can be found in [8]. For silhouette image $P(x,y)$, $m_{0,0}$ the moment of zero order represents the geometrical area of the image region and $m_{1,0}, m_{0,1}$ moment of first order represents the intensity moment about the y-axis and x-axis of the image respectively. The centroid $(x_c, y_c)$ gives the geometrical centre of the image region.

Suppose the size of the grid occupied by the object shape is N x N. The vector dimension to represent the density of object shape will be N-1. From the centroid we count the number of image pixels in the rings with defined equal width around the centroid. The number of image pixels in each ring is given as $X_i = (n_1, n_2, ......n_m)$ where m is the number of rings from the centroid.

The KDFPE is then applied. The Second-order Gaussian Kernel Density Estimator (SGKDE) is used. The SGKDE is given in [10] as

$$f(x) = \frac{1}{mh} \sum_{i=1}^{m} \frac{1}{\sqrt{2\pi}} e^{\frac{-1}{2}\left(\frac{x-X_i}{h}\right)^2} \quad (5)$$

The optimal bandwidth $h_o$ for each image shape is calculated using the following second order Gaussian plug-in formula [11]:

$$h_o = 1.059 m^{\frac{-1}{5}} s \quad (6)$$

where m is the number of rings from the centroid and $s$ is the sample standard deviation. The vector elements of the image are recalculated to make $f(x_i)$ becomes the image representation vector. Thus,

$$f(x_i) = \frac{1}{h_o \sqrt{2\pi}} e^{\frac{-1}{2}\left(\frac{X_i}{h_o}\right)^2} \quad (7)$$

For the purpose of illustration of this method, suppose object shape features are given on a grid as shown in Figure 1.

| 0,0 | **1,0** | **2,0** | **3,0** | **4,0** |
|---|---|---|---|---|
| 0,1 | **1,1** | **2,1** | **3,1** | **4,1** |
| **0,2** | **1,2** | **2,2** | **3,2** | **4,2** |
| 0,3 | 1,3 | 2,3 | **3,3** | **4,3** |
| 0,4 | 1,4 | 2,4 | **3,4** | **4,4** |

Figure 1. Segmented object shape

The red-bold indicate the "on" pixels, which belong to the image. The size of the grid occupied by the object shape is 5 x 5. The centroid calculated using Equations (2) and (3) is (3, 2), the centroid pixel is in blue. The first rectangle boundary in figure 1 is made up of the following pixels

$(2,1), (3,1), (4,1), (4,2), (4,3), (3,3), (2,3), (2,2)$

and there are seven "on" pixels that constitute our first element of the vector. The preliminary vector representation of object shape in Figure 1 is

(7, 8, 1)

The Kernel Density Estimator (KDE) is then applied. As mentioned earlier on the second-order Gaussian KDE is going to be used. The sample points (in our example) with the unknown distribution function are:

(7,8,1)

The second-order Gaussian KDE is given as:

$$f(x) = \frac{1}{3h} \sum_{i=1}^{3} \frac{1}{\sqrt{2\pi}} e^{\frac{-1}{2}\left(\frac{x-X_i}{h}\right)^2} \qquad (8)$$

The optimal bandwidth $h_o$ for each image shape is calculated. Then we recalculate the vector elements of the image to represent the image using the following

$$f(x_i) = \frac{1}{h_o \sqrt{2\pi}} e^{\frac{-1}{2}\left(\frac{X_i}{h_o}\right)^2} \qquad (9)$$

## 3. SIMILARITY MEASUREMENT

In order to measure the similarity of the images we used the cosine coefficient given in [12] as

$$s_{\cos} = \frac{\sum_{i=1}^{m} P_i Q_i}{\sqrt{\sum_{i=1}^{m} P_i^2} \sqrt{\sum_{i=1}^{m} Q_i^2}} \qquad (10)$$

The cosine coefficient, which is also called angular metric, is the normalized inner product of two vectors because it measures the angle between those vectors. The cosine coefficient has lower and upper bounds of 0 and 1 respectively. This makes it more suitable than Euclidean metric to establish comparison of results produced by two different image retrieval methods such as DHFP and KDFPE.

## 4. ACCURACY MEASUREMENT

The accuracy of an image retrieval system is generally measured by calculating recall, precision and effectiveness of the system. The following formulas were used [1]

$$recall = \frac{total\ relevent\ in\ the\ query\ results\ (A)}{total\ relevent\ in\ database\ (N)} \qquad (11)$$

$$precision = \frac{total\ relevent\ in\ the\ query\ results\ (A)}{total\ in\ the\ query\ results\ (A+C)} \qquad (12)$$

$$\mathit{effectiveness} = \begin{cases} \dfrac{A}{N} & \text{if } T > N \\ \dfrac{A}{T} & \text{if } T \leq N \end{cases}$$
(13)

In this case, $A$ is the number of relevant image objects retrieved, $B$ is the number of relevant image objects not retrieved, $C$ is the number of irrelevant image objects retrieved and $T$ is the user required number of relevant image retrieval.

## 5. EXPERIMENTATION

The main objective of the experimentation is to find the effectiveness of KDFPE method and to compare it with other representation methods. In this case a comparison is made with the DHFP. The cosine coefficient similarity measure is used in retrieval of similar image objects. An image database of 200 shop items shapes is created. Some of the image objects are rotated at 90, 180 and 270 degrees. The images that are rotated were not rotated lossless, meaning degradation of the image object occurred during rotation. The image objects were of different dimensions M x N or N x N where M and N are real numbers when they are brought to the system. The images that are used only have one image object with a homogeneous background. The image object shape of grid dimension 45 x 45 is segmented using the Chan and Vese active contour without edge [7]. All images are converted to gray scale images. They are then represented using KDFPE and DHFP. Each image was used as a query and the retrieval rate was measured using the Bull's Eye Performance (BEP), recall and precision performance. The average effectiveness is calculated for retrieval images in the database using queries of images captured using camera enabled devices. Matlab 7.6 was used to implement the system. Example of classes of shapes experimented with are given in Figure 2. In each class there are ten elements with some items rotated and scaled.

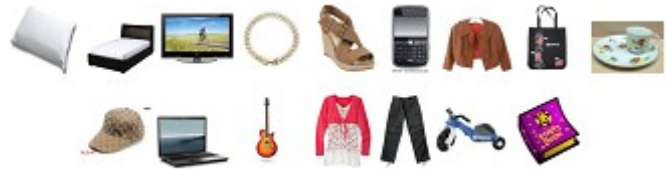

Figure 2. Examples of classes of items in database

## 6. RESULTS

Different types of televisions are in the collection of televisions in the database [13]. Some of the segmented televisions are shown in Figure 3 below. The KDFPE is capable of matching most of televisions despite errors in segmentation, noise and distortion of the television shapes because of transformations such as scaling and rotation.

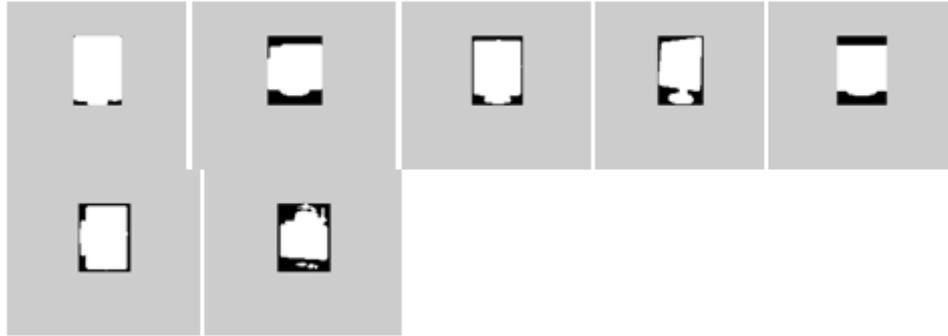

Figure 3. Segmented shapes that were considered similar by KDFPE

Figures 4a and 4b show how the system displays the query image and the retrieval results respectively. The system shows the user, the image used for querying the system and then the images considered similar to the query image.

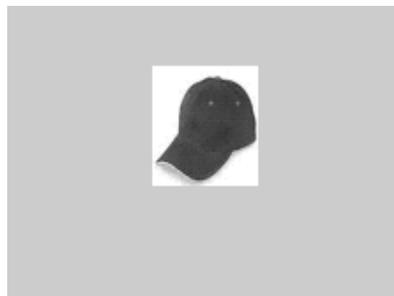

Figure 4 a. Query image

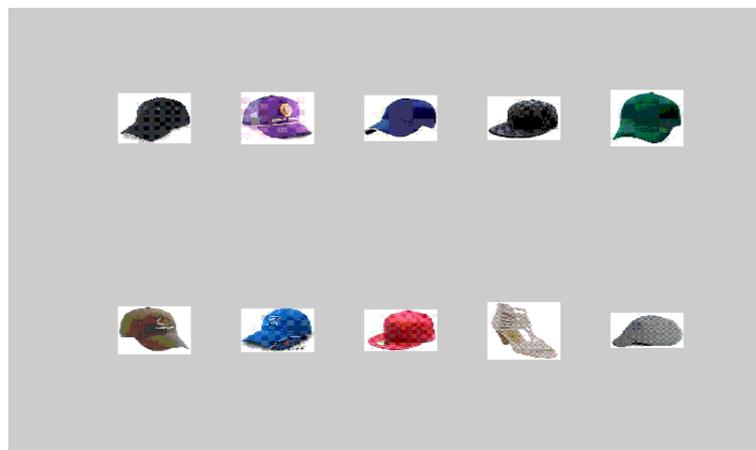

Figure 4b. Ten retrieval results of KDFPE (Figure 4a as query image)

Figure 5a shows the query image that serves as an input to the retrieval system. Figure 5b shows the images considered similar to the query image by the two methods. It can be seen that the query image is part of the retrieved images, indicating that it belongs to the database. In this sample of retrieval in Figure 5b, KDFPE has a 90% precision while DHFP has a 60% precision.

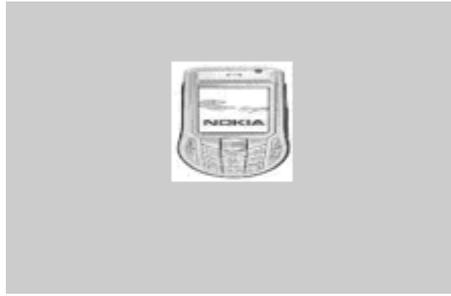

Figure 5a Query image

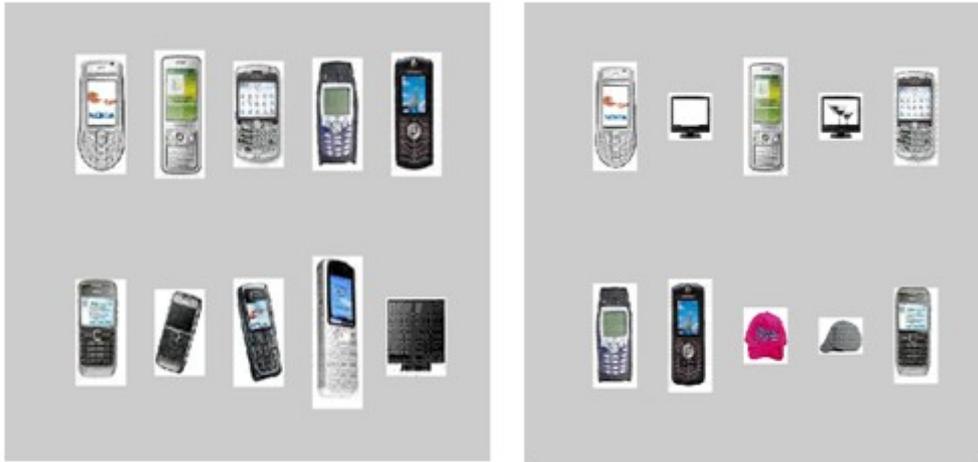

Figure 5b. Ten retrieval results of KDFPE on the left and DHFP on the right

The result in Figure 6 shows that KDFPE is better in retrieving images that are in the database using query images that belong to the database. It can be appreciated that the difference is not very substantial when it comes to querying the database with images in it. The Bull's Eye Performance (BEP) of DHFP is 91.18% while KDFPE is 92.40%.

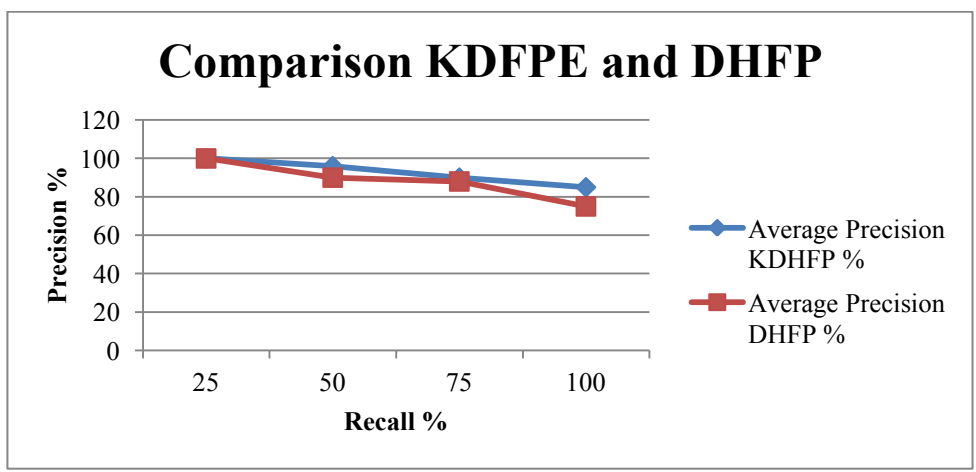

Figure 6. Comparison of KDFPE and DHFP using Recall-Precision Chart

The result in Figure 7 shows that KDFPE is better than the DHFP method when using a query that does not belong to the database. Figure 7 shows a very substantial difference in the effectiveness of the methods.

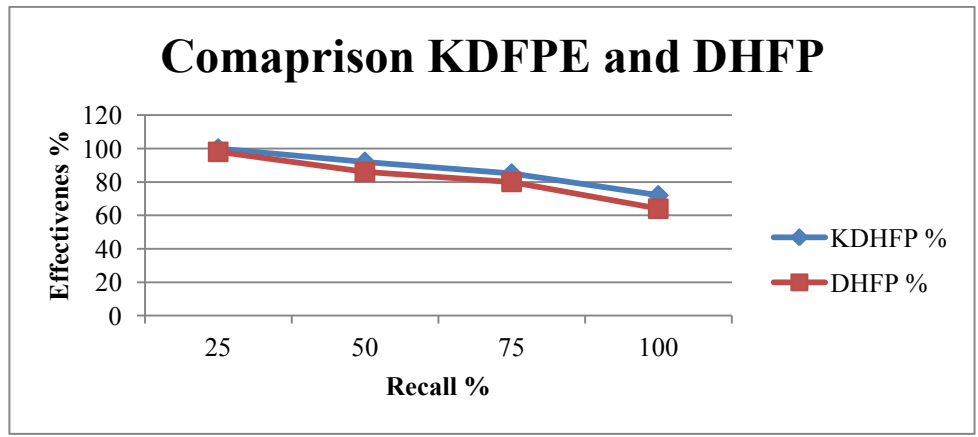

Figure 7. Comparison of effectiveness of the two methods using Recall-Precision Chart

## 7. CONCLUSION

From the results it can be concluded that KDFPE method of image object representation is able to differentiate similar object shapes just as human beings perceive image object shapes. The ability to calculate the optimal width seems to give KDFPE an advantage over DHFP method. Figure 7 shows a drastic reduction of the precision as the recall approaches 100% as compared to results in Figure 6. This is due to the way the image is captured. A 3-dimension item has so many angles that its image can be captured by a camera enabled device. This affects the retrieval rate of any object shape representation methods. The KDFPE method shows that it is capable of retrieving images similar to images captured by a camera enabled device as shown in figure 7. This characteristic is good in recommender systems because people are looking for images similar to their own captured images, but not necessarily in the database. In future we are going to use different methods of calculating optimal width to investigate if we may improve the effectiveness of KDFPE. Different kernel functions are going to be used in future to investigate if they have any effect in the retrieval rate. It is important to have a retrieval system that is capable of matching images that are in database with images that are not in the database. KDFPE is capable of overcoming errors in segmentation and is robust to segmentation noise.